\documentclass[runningheads]{llncs}

\usepackage[table]{xcolor}
\definecolor{bestcell}{RGB}{220,245,220}   
\definecolor{secondcell}{RGB}{255,240,200} 
\newcommand{\best}[1]{\cellcolor{bestcell}\textbf{#1}}
\newcommand{\second}[1]{\cellcolor{secondcell}#1}
\usepackage{siunitx}
\usepackage{subcaption}
\definecolor{deltacell}{RGB}{225,235,255} 
\newcommand{\deltaCell}[1]{\cellcolor{deltacell}\textbf{#1}}

\usepackage{eccv}

\usepackage{eccvabbrv}

\usepackage{graphicx}
\usepackage{booktabs}

\usepackage[accsupp]{axessibility}  

\usepackage{hyperref}

\begin{document}

\title{MoCA3D: Monocular 3D Bounding Box Prediction in the Image Plane} 


\author{Changwoo Jeon\inst{1,2} \and
Rishi Upadhyay\inst{1} \and
Achuta Kadambi\inst{1}}

\authorrunning{C. Jeon et al.}

\institute{University of California, Los Angeles \and
Yonsei University}

\maketitle

\begin{abstract}
    Monocular 3D object understanding has largely been cast as a 2D RoI-to-3D box lifting problem. However, emerging downstream applications require image-plane geometry (e.g., projected 3D box corners) which cannot be easily obtained without known intrinsics, a problem for object detection in the wild. We introduce \textbf{MoCA3D}, a \textbf{Mo}nocular, \textbf{C}lass-\textbf{A}gnostic \textbf{3D} model that predicts projected 3D bounding box corners and per-corner depths without requiring camera intrinsics at inference time. MoCA3D formulates pixel-space localization and depth assignment as dense prediction via corner heatmaps and depth maps. To evaluate image-plane geometric fidelity, we propose \textbf{Pixel-Aligned Geometry (PAG)}, which directly measures image-plane corner and depth consistency. Extensive experiments demonstrate that MoCA3D achieves state-of-the-art performance, improving image-plane corner PAG by 22.8\% while remaining comparable on 3D IoU, using up to $57\times$ fewer trainable parameters. Finally, we apply MoCA3D to downstream tasks which were previously impractical under unknown intrinsics, highlighting its utility beyond standard baseline models. Project page: \url{https://jeoncwcw.github.io/moca3d/}.

\end{abstract}

\section{Introduction}
\label{sec:intro}
Monocular 3D object understanding is a long-standing goal in computer vision, with applications in augmented reality~\cite{liu2019edge, park2008multiple, tekin2018real}, robotics~\cite{tremblay2018deep, labbe2020cosypose}, and visual servoing/control~\cite{choi2008real}.
Most monocular 3D approaches~\cite{zhou2019objects, li2020rtm3d, brazil2023omni3d, depthanything3} represent an object as a compact 3D entity in camera/world coordinates, such as an oriented cuboid or a 6D pose.
They are typically trained and evaluated in this 3D parameter space, and image-plane geometry is obtained only by projecting these 3D representations back to pixels when needed.

However, an increasing number of downstream applications (e.g., diffusion-based editing, controllable generation) consume geometry primarily in the image plane~\cite{wu2024neural, pandey2024diffusion, yang2025instadrive, sajnani2025geodiffuser, bhat2024loosecontrol, zhang2025perldiff, he2025morphosim, zhang2025instantrestore}.
In particular, they require accurate \emph{projected 3D bounding box corners} (often with assigned depths) to enforce perspective-consistent layout and spatial control.
This mismatch highlights a key limitation of existing monocular 3D pipelines: projected 3D corners are only actionable when camera intrinsics are known, which are often unavailable or inconsistent in real-world settings.

A natural alternative is to predict the required image-plane geometry directly from the RoI, as most instance lifting models attempt via direct regression of box or pose parameters. However, Yu \etal~\cite{yu2025boxdreamer} observe that direct corner regression is inherently sparse, providing limited gradient signals and potentially degrading performance; we therefore cast geometry recovery as a dense prediction task.

We propose \textbf{MoCA3D}, designed explicitly for \emph{image-plane-aligned} recovery.
Given a single RGB image and a tight oracle 2D bounding box, MoCA3D predicts projected 3D box corners and per-corner depths.
Importantly, MoCA3D does \emph{not} require camera intrinsics at inference time.
When intrinsics are available at evaluation, we can lift corners and depths into a shared 3D representation for fair comparison with monocular 3D detection baselines.

To realize accurate image-plane geometry from only a box, MoCA3D departs from standard RoI-to-vector regression in two key ways.
\textbf{(i) Leveraging transformer image priors.}
Instead of treating the box only as a crop, we encode the box into a spatial prior that conditions dense image tokens, and further inject box embeddings through cross-attention so that every spatial token can directly incorporate the instance specification.
\textbf{(ii) Dense prediction for corners and depth.}
To avoid sparse supervision, we cast recovery as dense prediction: MoCA3D produces eight corner heatmaps and per-corner depth maps, followed by differentiable soft-argmax extraction and depth sampling.
This design yields stable gradients across the RoI and encourages pixel-to-geometry alignment by construction.

A second challenge is evaluation: 3D metrics, such as 3D IoU, 3D corner distance, and pose error, can obscure reprojection errors that matter for downstream controls. We therefore introduce \textbf{Pixel-Aligned Geometry (PAG)}, a complementary metric suite that directly measures projected-corner error in the image plane (PAG$_{uv}$) and depth error at the corners (PAG$_d$).
Together with a 3D-space corner metric (NHD) and IoU$_{3D}$, PAG enables geometry-centric benchmarking in both the image plane and 3D space, by mapping diverse outputs to a unified evaluation protocol.
We show that MoCA3D substantially improves image-plane geometric fidelity over strong monocular 3D lifting baselines, while remaining lightweight and efficient. Finally, we demonstrate the practical utility of MoCA3D by integrating its outputs into controllable generation pipelines, where accurate reprojection and depth ordering are essential.

\paragraph{Contributions.}
Our main contributions are:
\begin {itemize}
    \item We propose \textbf{MoCA3D}, a box-conditioned, class-agnostic monocular 3D object geometry model that directly predicts image-plane 3D geometry as projected cuboid corners and per-corner depths, without requiring camera intrinsics at inference.
    \item We reformulate box-conditioned monocular 3D recovery as a dense prediction problem, using corner heatmaps and depth maps to provide richer supervision than direct RoI-to-3D regression.
    \item We introduce Pixel-Aligned Geometry (PAG) to evaluate reprojection and depth consistency in image space under an oracle-2D protocol, and show that MoCA3D delivers strong gains in image-plane geometry and efficiency.
\end{itemize}
\section{Related Work}

\subsubsection{Monocular 3D Object Detection.} Early monocular approaches were largely confined to specialized domains, such as autonomous driving~\cite{mousavian20173d, qin2021monogrnet} or indoor robotics~\cite{kim2020monocular}. 
However, these models struggle with ``in the wild'' scenarios due to variations in camera intrinsics and a limited vocabulary of object classes. Cube R-CNN~\cite{brazil2023omni3d} addressed these limitations by introducing a camera-agnostic virtual depth space, enabling the first large-scale universal 3D detector trained on the Omni3D benchmark.

Concurrently, \emph{keypoint-based detection}~\cite{law2018cornernet, zhou2019objects} recasts object detection as keypoint estimation by predicting sparse landmarks such as box corners or centers as heatmaps and regressing box parameters.
This paradigm has also been adopted for monocular 3D detection: CenterNet-style designs regress 3D parameters (depth, dimensions, and orientation) at detected centers~\cite{wang2021centernet3d, li2021monocular, liu2020smoke}, while RTM3D~\cite{li2020rtm3d} predicts perspective keypoints of a 3D bounding box in the image plane and recovers 3D boxes via geometric constraints.
In contrast to predicting sparse keypoints and regressing a compact 3D parameter vector, MoCA3D directly predicts projected 3D box corners with per-corner depths via dense heatmap and depth map prediction.

\subsubsection{Coordinate Based Representations.} A parallel line of research leverages Normalized Object Coordinate Space (NOCS)~\cite{Wang_2019_CVPR} to bridge the gap between 2D images and 3D geometry. While early NOCS methods were restricted to small-scale indoor datasets, recent works like OmniNOCS~\cite{krishnan2024omninocs} and SOCS~\cite{wan2023socs} scale this representation to hundreds of thousands of instances across diverse environments. By predicting dense 3D coordinates and utilizing Perspective-n-Point (PnP) solvers, these methods provide more granular shape information than traditional cuboid-regressing frameworks.
Compared to these pipelines, MoCA3D predicts pixel-aligned corner-depth pairs directly in the image plane, rather than estimating object coordinates and solving for pose.

\subsubsection{Open-World and Foundation Models.} Recent advances have moved toward open-vocabulary 3D detection. Rather than relying solely on 3D supervision, methods such as DetAny3D~\cite{zhang2025detect} and 3D-MOOD~\cite{yang20253d} leverage the rich spatial priors of 2D foundation models (e.g., SAM~\cite{ravi2024sam2}, DINO~\cite{simeoni2025dinov3}). These ``lifting'' strategies allow for zero-shot generalization to novel categories and unconstrained camera configurations, representing the current state-of-the-art for 3D perception in the wild. MoCA3D is complementary and similarly targets open-world settings via class-agnostic, box-conditioned geometry recovery.

\section{Method}
\label{sec:method}
\subsection{Overview}
\begin{figure*}[t]
  \centering
  \includegraphics[width=\textwidth]{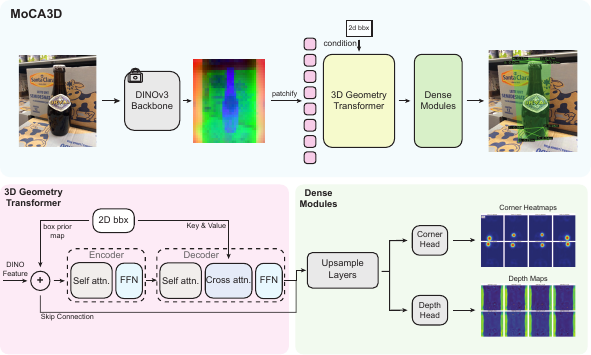}
  \caption{\textbf{MoCA3D architecture.} Given an RGB image and a tight oracle 2D bounding box, MoCA3D uses a frozen DINOv3 backbone and a box-conditioned 3D Geometry Transformer with dense modules to predict eight corner heatmaps and per-corner depth maps, yielding pixel-aligned projected 3D box corners and depths.
  }
  \label{fig:moca3d}
\end{figure*}

Given a single RGB image $I \in \mathbb{R}^{H\times W\times 3}$ and a 2D bounding box $\mathbf{b} = (x_1, y_1, x_2, y_2)$ around an object instance, MoCA3D predicts the output as $\hat{\mathcal{P}} = \{(\hat{\mathbf{p}}_i, \hat{d}_i)\}_{i=1}^{8}$ where $\hat{\mathbf{p}}_i=(\hat{u}_i, \hat{v}_i)$ denotes the image-plane box corner location (in pixels) and $\hat{d}_i$ is the corresponding depth. As illustrated in Fig.~\ref{fig:moca3d}, MoCA3D consists of three stages. \textbf{(i) Backbone feature extraction}, where a frozen ViT foundation model produces a feature map $\mathbf{F}^{\text{dino}}\in\mathbb{R}^{C_0\times h_d\times w_d}$ from $I$. We use DINOv3~\cite{simeoni2025dinov3} to obtain geometry-aware embeddings. \textbf{(ii) Box-conditioned 3D Geometry Transformer}, where the 2D box $\mathbf{b}$ is encoded into a spatial box prior map that conditions the encoder and box embeddings for the decoder, enabling the transformer to refine image tokens within the object extent. \textbf{(iii) Dense geometry heads}, where lightweight upsampling layers feed two heads: a corner head predicting eight corner heatmaps and a depth head predicting per-corner depth maps.

\subsection{3D Geometry Transformer}
\label{sec:geoformer}

We build a box-conditioned transformer (Fig.~\ref{fig:moca3d} bottom left) following the canonical formulation of Transformer~\cite{vaswani2017attention} and the DETR-style encoder-decoder design for vision~\cite{carion2020end}. 

\noindent\textbf{Box prior map (relative-coordinate encoding) and Encoder.} Starting from the frozen DINOv3~\cite{simeoni2025dinov3} features $\mathbf{F}^{\text{dino}}\in\mathbb{R}^{C_0\times h_d\times w_d}$, we map them to the transformer hidden size $C$ and inject an explicit per-pixel box prior map $\mathbf{M}(\mathbf{b})\in\mathbb{R}^{4\times h_d\times w_d}$ to condition dense tokens on the target instance. Specifically, for each spatial location $(x,y)$, we encode its position relative to the box using four channels. Center-normalized offsets $(d_x,d_y)$ which indicate whether a pixel lies left/right or above/below the box center and box-normalized coordinates $(u,v)$ that measure the pixel's relative location w.r.t.\ the top-left corner. 
We project the prior map to the hidden dimension and inject it additively with a learned gate:
\begin{equation}
\mathbf{E}_{\text{prior}}=\psi(\mathbf{M}(\mathbf{b}))\in\mathbb{R}^{C\times h_d\times w_d},\qquad
\alpha=\sigma(\gamma),\qquad
\tilde{\mathbf{F}}=\mathbf{F}+\alpha\,\mathbf{E}_{\text{prior}},
\label{eq:priorfuse}
\end{equation}
where $\psi(\cdot)$ is a learnable projection and $\gamma$ is a learnable scalar logit. 
We then add a standard 2D positional encoding and flatten the feature grid into a sequence of $N=h_dw_d$ tokens. An $L_e$-layer Transformer encoder contextualizes these tokens with multi-head self-attention and feed-forward networks with residual connections. 

\noindent\textbf{Decoder and box embeddings.} Unlike detection transformers~\cite{liu2022dab, zhu2020deformable, carion2020end, meng2021conditional, li2022dn} that use a small set of object queries to retrieve information from image features, our goal is to refine dense image tokens for pixel-aligned prediction. We therefore reverse the usual conditioning direction: the encoded image tokens act as queries, while the box embeddings provide keys/values, so that every spatial token can directly incorporate the oracle box signal. Concretely, we encode the oracle box $\mathbf{b}$ into a small set of box embeddings $\mathbf{B}\in\mathbb{R}^{C\times N_q}$ (we use $N_q{=}9$), then apply an $L_d$-layer Transformer decoder that utilizes (i) self-attention over the image tokens, (ii) cross-attention where image tokens attend to the box embeddings as keys/values, and (iii) a feed-forward network, all with residual connections.

\subsection{Dense Modules}
\label{sec:dense_modules}

Given the decoder output tokens, we reshape them back to the feature grid to obtain a 2D feature map $\mathbf{D}\in\mathbb{R}^{C\times h_d\times w_d}$. We fuse $\mathbf{D}$ with the box-conditioned backbone feature $\tilde{\mathbf{F}}$ (Eq.~\eqref{eq:priorfuse}) using a lightweight fusion block,  $\mathbf{G}=\phi_{\text{fuse}}([\mathbf{D};\tilde{\mathbf{F}}])\in\mathbb{R}^{C\times h_d\times w_d}$, where $[\cdot;\cdot]$ denotes channel-wise concatenation.
We then progressively upsample $\mathbf{G}$ with two lightweight stages to obtain a higher-resolution feature map $\mathbf{U}\in\mathbb{R}^{C/4\times h_o\times w_o}$ for dense prediction. In our implementation, the output resolution is $h_o=h_d\times4$, and the channel dimension is reduced from $C$ to $C/4$ for efficiency.

\noindent\textbf{Dense prediction heads.}
From $\mathbf{U}$, we predict (i) eight corner heatmaps and (ii) per-corner depth maps using two small convolutional heads:
\begin{equation}
\mathbf{H}=\phi_{\text{hm}}(\mathbf{U})\in[0,1]^{8\times h_o\times w_o},\qquad
\mathbf{Z}=\phi_{\text{dep}}(\mathbf{U})\in\mathbb{R}_{+}^{8\times h_o\times w_o},
\label{eq:heads}
\end{equation}
where $\mathbf{H}$ and $\mathbf{Z}$ denote heatmap and depth map, and $\phi_{hm}$ and $\phi_{dep}$ denote corner head and depth head. $\phi_{\text{hm}}$ ends with a sigmoid activation to ensure a bounded heatmap interpretation consistent with the target heatmap values defined in our loss (Sec.~\ref{sec:loss}), facilitating stable alignment between predictions and supervision. In contrast, $\phi_{\text{dep}}$ ends with a softplus~\cite{glorot2011deep} to enforce non-negative depths.
We initialize the heatmap head bias to encourage low initial activations, which helps stabilize early-stage training.

\noindent\textbf{Soft-argmax~\cite{luvizon2019human} corner extraction and depth sampling.}
We convert each corner heatmap $\mathbf{H}_i$ into a continuous coordinate $\hat{\mathbf{p}}_i\in\mathbb{R}^2$ using a differentiable soft-argmax function.
Concretely, soft-argmax first forms a spatial probability distribution by applying a softmax over all locations with an inverse-temperature $\beta$:
\begin{equation}
\pi_i(x,y)=\frac{\exp(\beta\,\mathbf{H}_i(x,y))}{\sum_{x',y'}\exp(\beta\,\mathbf{H}_i(x',y'))},
\label{eq:softargmax}
\end{equation}
and returns the expected coordinate under $\pi_i$:
\begin{equation}
\hat{u}_i=\sum_{x,y} x\,\pi_i(x,y),\qquad
\hat{v}_i=\sum_{x,y} y\,\pi_i(x,y),\qquad
\hat{\mathbf{p}}_i=(\hat{u}_i,\hat{v}_i).
\label{eq:softargmax_expect}
\end{equation}
The larger $\beta$ approaches the hard argmax (more peak-like responses), while smaller $\beta$ yields smoother averaging, allowing sub-pixel localization with stable gradients.
Finally, we sample per-corner depths from $\mathbf{Z}$ at the predicted corner locations using bilinear interpolation.
\begin{equation}
\hat{d}_i=\mathrm{Sample}\!\left(\mathbf{Z}_i,\hat{\mathbf{p}}_i\right),\qquad i=1,\dots,8.
\label{eq:depthsample}
\end{equation}
For consistency with our evaluation setting, the coordinates are scaled from the heatmap grid back to the image resolution, yielding the final predictions. $\hat{\mathcal{P}}=\{(\hat{\mathbf{p}}_i,\hat{d}_i)\}_{i=1}^{8}$.
\subsection{Loss}
\label{sec:loss}

\begin{figure*}[t]
  \centering
  \includegraphics[height=5.5cm, trim=0 10 0 0,clip]{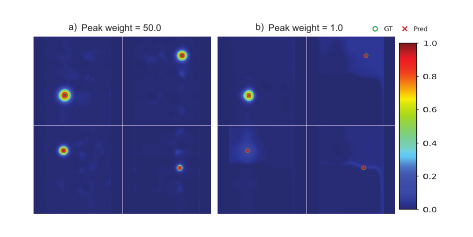}
  \caption{\textbf{Heatmap Comparison.} Predicted corner heatmaps with (a) peak weight $=50.0$ and (b) $=1.0$ (uniform). Larger peak weight sharpens and localizes responses near GT corners, improving soft-argmax stability, while uniform weighting yields flatter heatmaps.
  }
  \label{fig:heatmap_comp}
\end{figure*}

We supervise MoCA3D with a peak-weighted heatmap regression loss and a coordinate-level refinement loss for corner localization, together with a confidence-weighted depth loss. Let $\{\mathbf{p}_i\}_{i=1}^{8}$ and $\{d_i\}_{i=1}^{8}$ denote the ground-truth projected corners and depths.

\noindent\textbf{Target heatmap generation.}
By placing a Gaussian-like peak around each ground-truth corner, we construct a soft target heatmap $\mathbf{W}\in[0,1]^{8\times h_o\times w_o}$:

\begin{equation}
\mathbf{W}_i(x,y)=\exp\!\left(-\frac{\|(x,y)-\mathbf{p}_i\|_2^2}{2\sigma_i^2}\right),
\label{eq:target_heatmap}
\end{equation}
where $2\sigma_i^2$ is set adaptively based on the object size, following Yu \etal~\cite{yu2025boxdreamer} (we set $2\sigma_i^2$  as the squared one-fifth of pixel distance between object's 2D center and each corner). 

\noindent\textbf{Peak-weighted heatmap loss (coarse).}
A naive heatmap regression treats background and peak pixels similarly; however, peak supervision is inherently sparse and can be overwhelmed by the vast background, leading to diffuse responses that are less informative for downstream coordinate extraction. To explicitly prioritize peak alignment, we upweight pixels whose target heatmap value exceeds a threshold $\tau$:
\begin{equation}
\mathbf{A}_i(x,y)=
\begin{cases}
\lambda, & \mathbf{W}_i(x,y)>\tau,\\
1, & \text{otherwise},
\end{cases}
\label{eq:peak_weight}
\end{equation}
where $\lambda>1$ is a peak emphasis factor. We then apply robust regression loss:
\begin{equation}
\mathcal{L}_{\text{coarse}}=
\frac{\sum_{i,x,y}\mathbf{A}_i(x,y)\cdot
\mathrm{SmoothL1}\!\left(\mathbf{H}_i(x,y)-\mathbf{W}_i(x,y)\right)}
{\sum_{i,x,y}\mathbf{A}_i(x,y)+\epsilon}.
\label{eq:l_coarse}
\end{equation}
As visualized in Fig.~\ref{fig:heatmap_comp}, increasing the peak weight concentrates gradients around the true corner locations, producing sharper and more localized maxima, while uniform weighting yields flatter and less reliable heatmaps.

\noindent\textbf{Coordinate refinement loss (fine).}
To encourage accurate sub-pixel localization, we apply a coordinate-level loss on the soft-argmax outputs:
\begin{equation}
\mathcal{L}_{\text{fine}}=\frac{1}{8}\sum_{i=1}^{8}
\mathrm{SmoothL1}\!\left(\hat{\mathbf{p}}_i-\mathbf{p}_i\right),
\label{eq:l_fine}
\end{equation}
Importantly, $\mathcal{L}_{\text{coarse}}$ and soft-argmax make $\mathcal{L}_{\text{fine}}$ more effective: as the peak-weighted coarse loss sharpens $H_i$, the soft-argmax concentrates around the true mode, producing sub-pixel coordinates and cleaner gradients for refinement.

\noindent\textbf{Confidence-weighted depth loss (pixel-aligned depth supervision).}
Following Cube R-CNN~\cite{brazil2023omni3d}, MoCA3D regresses virtual depth to alleviate scale ambiguity, and supervises per-corner depth with a confidence-weighted loss in which the predicted heatmaps act as per-pixel reliability weights. Let $\bar{d}_i\in\mathbb{R}^{h_o\times w_o}$ denote the ground-truth for that corner (broadcast to the heatmap grid). We define:
\begin{equation}
    \mathcal{L}_{\text{depth}}=
\frac{\sum_{i,x,y}\mathbf{H}_i(x,y)\cdot
\mathrm{SmoothL1}\!\left(\mathbf{Z}_i(x,y)-\bar{d}_i\right)}
{\sum_{i,x,y}\mathbf{H}_i(x,y)+\epsilon}.
\label{eq:l_depth}
\end{equation}
This formulation explicitly couples localization and depth estimation: pixel-aligned heatmaps indicate where each corner is likely to lie, thereby focusing depth supervision on spatial regions that are consistent with predicted corner location. Consequently, more reliable heatmaps (Fig.~\ref{fig:heatmap_comp}) not only improve $\hat{\mathbf{p}}_i$, but also provide more accurate and stable sampling for per-corner depth.

\noindent\textbf{Total loss and warm-up.}
The final objective is:
\begin{equation}
\mathcal{L} = 
w_{\text{coarse}}\mathcal{L}_{\text{coarse}} +
w_{\text{fine}}\mathcal{L}_{\text{fine}} +
w_{\text{depth}}\mathcal{L}_{\text{depth}}.
\label{eq:total_loss}
\end{equation}
We use a warm-up schedule: we first optimize only the coarse heatmap loss, then enable the refinement and depth terms and linearly re-balance the weights over training ($w_{\text{coarse}}:50{\rightarrow}1$, $w_{\text{fine}}:0{\rightarrow}2$, $w_{\text{depth}}:0{\rightarrow}5$).

\subsection{MoCA3D-Cube}
To demonstrate MoCA3D's adaptability to standard 3D bounding box prediction, we design \textbf{MoCA3D-Cube}, a lightweight 3D bounding box adapter.
MoCA3D-Cube takes as input (i) MoCA3D's projected corner coordinates $\hat{\mathbf{p}}$ and depth outputs $\hat{d}$ and (ii) a RoI-aligned embedding pooled from the decoder feature map. These features are concatenated and passed to a compact two-layer Cube MLP that regresses a parametric 3D bounding box, including center, size, and rotation with uncertainty, conditioned on camera intrinsics $K$. For training and supervision, we follow the Cube R-CNN~\cite{brazil2023omni3d} protocol for 3D box parameterization and losses. We initialize from a pretrained MoCA3D and jointly fine-tune the adapter with MoCA3D while keeping the backbone frozen.

\begin{figure*}[!t]
  \centering
  \includegraphics[width=\textwidth]{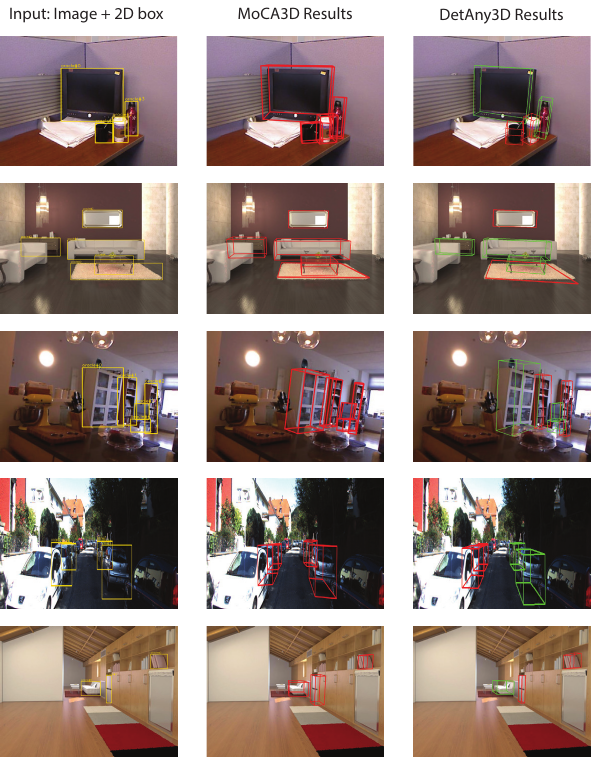}
  \caption{\textbf{Qualitative Results.}
    MoCA3D vs.\ DetAny3D predictions under oracle 2D boxes on samples from the KITTI, Omni3D, and Hypersim datasets. Detections in green have a 3D IoU of less than 0.1, making them low-quality detections.}
  \label{fig:qual}
\end{figure*}

\section{Experiments}
\label{sec:experiments}

\subsection{Experimental Settings}
\label{subsec: setting}
\subsubsection{Datasets and Baselines.} We evaluate \textbf{MoCA3D} on the Omni3D benchmark~\cite{brazil2023omni3d} that provides diverse spatial coverage across multiple domains (urban, indoor, and general) by integrating datasets and annotations from nuScenes~\cite{nuscenes}, KITTI~\cite{kitti}, SUN RGB-D~\cite{sunrgbd}, ARKitScenes~\cite{arkitscenes}, Hypersim~\cite{hypersim} and Objectron~\cite{objectron}. However, Omni3D contains instances with missing 2D boxes; we complete them using SAM2 with projected 3D corners as prompts (see supplementary material for preprocessing details).

We compare \textbf{MoCA3D} against Cube R-CNN~\cite{brazil2023omni3d}, OVMono3D-Lift$^{\ast}$~\cite{yao2024open}, and DetAny3D~\cite{zhang2025detect}. Although these methods are developed for monocular 3D detection, we benchmark them under a unified protocol that removes detector confounds: given the box as input, and each method performs \emph{box-conditioned 3D lifting} to predict a 3D box. This enables an apples-to-apples comparison with MoCA3D; both can be mapped into a shared representation using camera intrinsics $K$ for evaluation. Cube R-CNN operates in a closed-set setting, whereas OVMono3D and DetAny3D are open vocabulary models. To further isolate the effect of category priors, we additionally train a Cube R-CNN$^{\ast}$ variant by removing class priors while keeping the remaining training and inference settings unchanged.
\subsubsection{Implementation Details.} We use DINOv3 ViT-L/16~\cite{simeoni2025dinov3} as the frozen image backbone and train on Omni3D~\cite{brazil2023omni3d} with oracle tight 2D boxes. 
Inputs are resized to $512\times512$ using aspect-ratio preserving \emph{letterboxing} (padding), and the same transform is applied to ground-truth 2D boxes and projected corners.
Training is performed for 120 epochs on 3 NVIDIA RTX 3090 GPUs (50{,}000 instances per epoch; \mbox{$\sim$60 GPU-hours} total). For efficiency, we pre-extract and cache backbone features, which requires \mbox{$\sim$4 GPU-hours.} 
We optimize with AdamW~\cite{loshchilov2017decoupled} using a learning rate of $2\times10^{-4}$ and a cosine schedule~\cite{loshchilov2016sgdr} with a 5-epoch warm-up. We apply standard augmentations including box jittering, feature dropout, horizontal flipping, and rotation; further details are provided in the supplementary material.

\subsection{Metrics}
\label{subsec: metric} 
Following CatFree3D~\cite{bian2025catfree3d}, we report Normalized Hungarian Distance (NHD). NHD computes an optimal one-to-one correspondence between the eight corners using Hungarian matching and sums their Euclidean distances, normalized by the ground-truth box diagonal:
\begin{equation}
\mathrm{NHD}(M_{\text{gt}}, M_{\text{pred}})=\frac{1}{d_{\text{gt}}}\sum_{i=1}^{8}\left\lVert \mathbf{a}_i-\mathbf{b}_{\pi(i)}\right\rVert_2,
\end{equation}
where $\pi$ is the optimal assignment and $d_{\text{gt}}$ denotes the diagonal length of the ground-truth 3D box $M_{\text{gt}}$.

To evaluate image-plane geometry, we additionally report Pixel-Aligned Geometry (PAG) metrics that operate on the predicted corner-depth pairs. Recall from Sec.~\ref{sec:method} that MoCA3D predicts $\hat{\mathcal{P}}=\{(\hat{\mathbf{p}}_i,\hat d_i)\}_{i=1}^{8}$, where $\hat{\mathbf{p}}_i\in\mathbb{R}^2$ is the image-plane corner location and $\hat d_i$ is its depth; ground truth is denoted by $\mathcal{P}=\{(\mathbf{p}_i,d_i)\}_{i=1}^{8}$.
We define the image-plane corner error (in pixels) as:
\begin{equation}
\mathrm{PAG}_{uv}=\frac{1}{8}\sum_{i=1}^{8}\left\lVert \hat{\mathbf{p}}_i-\mathbf{p}_i\right\rVert_2,
\end{equation}
and the relative depth error (\%) as:
\begin{equation}
\mathrm{PAG}_{d}=\frac{100}{8}\sum_{i=1}^{8}\frac{\lvert \hat d_i-d_i\rvert}{d_i},
\end{equation}
Unless otherwise specified, we compute PAG on images resized to $512\times512$ for consistent comparison across methods. Together, NHD and IoU$_{3D}$ evaluate geometric accuracy in 3D bounding box space, while PAG measures pixel-aligned geometry via image-plane corner error (PAG$_{uv}$) and relative depth error (PAG$_d$). IoU$_{3D}$ is implemented using PyTorch3D~\cite{lassner2020pulsar}.

\subsection{Comparisons with Baselines}

\begin{table}[t]
\centering
\scriptsize
\setlength{\tabcolsep}{2.5pt}
\renewcommand{\arraystretch}{0.95}

\resizebox{\linewidth}{!}{
\begin{tabular}{lcccccc}
\toprule
\multicolumn{7}{c}{\textbf{PAG$_{uv}$ (px) $\downarrow$}}\\
\midrule
Model & kitti & nuscenes & objectron & arkitscenes & sunrgbd & hypersim \\
\midrule
Cube R-CNN        & 10.10 & 11.73 & 23.57 & 22.47 & 23.17 & 13.92 \\
Cube R-CNN$^{\ast}$       & 10.38 & 11.86 & 31.28 & 23.75 & 24.06 & 14.17 \\
OVMono3D-LIFT$^{\ast}$    & 10.29 & 11.82 & \second{23.55} & 22.55 & 22.36 & 13.02 \\
DetAny3D          & \second{4.90} & \second{6.75} & 27.41 & \second{17.15} & \second{21.06} & \second{10.71} \\
MoCA3D            & \best{3.12} & \best{5.63} & \best{18.60} & \best{14.90} & \best{16.99} & \best{9.56} \\
\bottomrule
\end{tabular}
}
\resizebox{\linewidth}{!}{
\begin{tabular}{lcccccc}
\toprule
\multicolumn{7}{c}{\textbf{PAG$_d$ (\%) $\downarrow$}}\\
\midrule
Model & kitti & nuscenes & objectron & arkitscenes & sunrgbd & hypersim \\
\midrule
Cube R-CNN        & 12.63 & 12.05 & 15.03 & 13.22 & 13.92 & 20.66 \\
Cube R-CNN$^{\ast}$       & 13.85 & 13.04 & 33.18 & 14.92 & 16.61 & 22.81 \\
OVMono3D-LIFT$^{\ast}$    & 13.20 & 12.45 & \second{12.68} & 12.60 & 14.61 & 19.94 \\
DetAny3D          & \best{4.82} & \best{6.07} & \best{12.55} & \second{8.87} & \best{8.71} & \best{16.81} \\
MoCA3D            & \second{5.04} & \second{7.17} & 14.59 & \best{8.78} & \second{10.08} & \second{19.57} \\
\bottomrule
\end{tabular}
}

\caption{\textbf{Image-plane geometry metrics: MoCA3D outperforms or is comparable to prior models across all datasets.} All results were computed with oracle 2D bounding boxes and ground truth intrinsics were used to calculate pixel-space coordinates for models other than MoCA3D. Green indicates the best and orange indicates the second-best results. Cube R-CNN$^\ast$ is trained in a class-agnostic setting.}
\label{tab:oracle2d_pag_combined}
\end{table}

\begin{table*}[t]
\centering
\scriptsize
\setlength{\tabcolsep}{2.5pt}
\renewcommand{\arraystretch}{0.95}

\resizebox{\linewidth}{!}{
\begin{tabular}{lcccccc}
\toprule
\multicolumn{7}{c}{\textbf{NHD $\downarrow$}}\\
\midrule
Model & kitti & nuscenes & objectron & arkitscenes & sunrgbd & hypersim \\
\midrule
Cube R-CNN        & 0.2811 & 0.3067 & 0.2462 & 0.3590 & 0.4466 & 1.0603 \\
Cube R-CNN$^{\ast}$       & 0.3504 & 0.3758 & 0.5282 & 0.4074 & 0.5642 & 1.1875 \\
OVMono3D-LIFT$^{\ast}$    & 0.3177 & 0.3458 & \second{0.2132} & 0.3527 & 0.4864 & \second{1.0216} \\
DetAny3D          & \second{0.1911} & \best{0.2428} & \best{0.2107} & \second{0.2593} & \best{0.3026} & \best{0.8970} \\
MoCA3D            & \best{0.1902} & \second{0.3002} & 0.2519 & \best{0.2529} & \second{0.3381} & 1.0406 \\
\bottomrule
\end{tabular}
}
\label{tab:oracle2d_nhd_compact}

\resizebox{\linewidth}{!}{
\begin{tabular}{lcccccc}
\toprule
\multicolumn{7}{c}{\textbf{IoU$_{3D}$ $\uparrow$}}\\
\midrule
Model & kitti & nuscenes & objectron & arkitscenes & sunrgbd & hypersim \\
\midrule
Cube R-CNN        & 0.4387 & \second{0.4118} & 0.3501 & 0.3135 & \second{0.2908} & 0.0895 \\
Cube R-CNN$^{\ast}$       & 0.3738 & 0.3569 & 0.2748 & 0.2534 & 0.2260 & 0.0769 \\
OVMono3D-LIFT$^{\ast}$    & 0.3812 & 0.3755 & \best{0.3706} & 0.3060 & 0.2419 & \second{0.0899} \\
DetAny3D          & \best{0.4834} & \best{0.4311} & \second{0.3587} & \best{0.3826} & \best{0.3272} & \best{0.1079} \\
MoCA3D            & \second{0.4735} & 0.3962 & 0.3015 & 0.3180 & 0.2753 & 0.0713 \\
MoCA3D-Cube       & 0.4549 & 0.3993 & 0.3163 & \second{0.3435} & 0.2455 & \second{0.0899} \\
\bottomrule
\end{tabular}
}
\label{tab:oracle2d_iou_compact}

\caption{\textbf{3D camera space metrics: MoCA3D is competitive with prior work such as DetAny3D despite a significantly smaller parameter count and compute requirements.} Metrics were computed using ground truth camera intrinsics and oracle 2D bounding boxes. Green indicates the best and orange indicates the second-best results.}
\label{tab:oracle2d_box_combined}
\end{table*}

\noindent\textbf{Image-plane Evaluations}
Table~\ref{tab:oracle2d_pag_combined} evaluates pixel-aligned geometry under the oracle-2D protocol. Across all six domains, MoCA3D achieves the lowest PAG$_{uv}$, indicating the most accurate projected corner localization. This suggests that MoCA3D yields stronger pixel-level supervision than detection-style lifting pipelines. Aggregated over Omni3D, MoCA3D improves PAG$_{uv}$ by 22.8.\%.
MoCA3D ranks second overall on PAG$_d$ across the six domains, achieving the best on ARKitScenes~\cite{arkitscenes} (8.78\%) and the second on KITTI~\cite{kitti}, nuScenes~\cite{nuscenes}, SUN RGB-D~\cite{sunrgbd}, and Hypersim~\cite{hypersim}, while remaining competitive on Objectron~\cite{objectron}. 

\noindent\textbf{3D Box Evaluation}
Table~\ref{tab:oracle2d_box_combined} reports NHD and IoU$_{3D}$ under oracle-2D protocol.
On NHD, MoCA3D achieves the best performance on KITTI~\cite{kitti} and ARKitScenes~\cite{arkitscenes} and is comparable to the strongest 3D lifting baseline (DetAny3D~\cite{zhang2025detect}) across domains, indicating accurate 3D corner geometry even without explicitly enforcing a cuboid parameterization. 

To compute IoU$_{3D}$, we rectify MoCA3D's unprojected corners into the closest valid cuboid using a Kabsch-based~\cite{kabsch1976solution} rigid alignment. Specifically, we unproject $(\hat{\mathbf{p}}_i, \hat d_i)$ with $K$ to obtain 3D corner points, then align them to a canonical cuboid template via a Procrustes/Kabsch solve and recover per-axis scale to form an oriented cuboid. This produces a cuboid with a consistent vertex topology, allowing direct evaluation with PyTorch3D~\cite{lassner2020pulsar} IoU$_{3D}$. MoCA3D achieves the second-best performance on KITTI~\cite{kitti}, and is comparable to Cube R-CNN and OVMono3D-Lift$^\ast$ across the remaining datasets.

Finally, MoCA3D-Cube explicitly adapts image-plane geometry to a parametric box representation.
As a result, it consistently improves IoU$_{3D}$ over MoCA3D except on SUNRGBD~\cite{sunrgbd} and KITTI~\cite{kitti} and attains second-best performance on ARKitScenes~\cite{arkitscenes} and Hypersim~\cite{hypersim}, demonstrating that MoCA3D's representation can be readily mapped to conventional 3D detection outputs when box-level evaluation is desired.

\begin{figure*}[t!]
  \centering
  \makebox[\textwidth][c]{%
    \includegraphics[width=\textwidth]{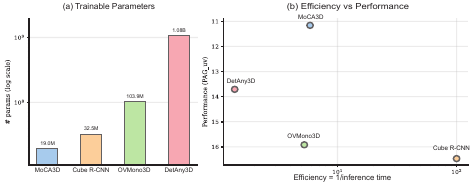}%
  }
  \caption{\textbf{Efficiency of MoCA3D.}
    We compare (a) trainable parameters and (b) trade-off between efficiency and performance (PAG$_{uv}$). Efficiency is defined as the inverse of end-to-end inference time per example on CV-Bench~\cite{tong2024cambrian1}.}
  \label{fig:efficiency}
\end{figure*}

\noindent\textbf{Efficiency of MoCA3D}
MoCA3D is lightweight (\SI{19.0}{M} trainable parameters, 57$\times$ smaller than DetAny3D), runs efficiently at inference time (0.14 s/ex) at CV-Bench~\cite{tong2024cambrian1}.
Moreover, in Fig.~\ref{fig:efficiency}(b), MoCA3D attains the lowest PAG$_{uv}$ while remaining highly efficient, demonstrating a favorable efficiency-performance trade-off.

\subsection{Ablations}
\label{ablation}

\begin{table}[t]
  \centering
  \small
  \setlength{\tabcolsep}{6.0pt}
  \renewcommand{\arraystretch}{1.15}

  \resizebox{\linewidth}{!}{%
  \begin{tabular}{l
                  S[table-format=2.2]
                  S[table-format=2.2]
                  S[table-format=1.4]
                  S[table-format=2.1]
                  S[table-format=5.1]}
    \toprule
    \textbf{Method} &
    {\textbf{PAG$_{uv}$} $\downarrow$ (px)} &
    {\textbf{PAG$_{d}$} $\downarrow$ (\%)} &
    {\textbf{IoU$_{3D}$} $\uparrow$} &
    {\textbf{\#Params} (M)} &
    {\textbf{Train} (GPU-hrs)} \\
    \midrule

    MoCA3D (baseline)  & \textbf{16.05} & \textbf{10.83} & \textbf{0.3768} & 19.0 & 27.0 \\
    \midrule

    MoCA3D w/ DA       & 16.14 & 10.92 & 0.3682 & 19.8 & {--} \\
    \quad $\Delta$ vs.\ MoCA3D
                       & \deltaCell{+0.09} & \deltaCell{+0.09} & \deltaCell{-0.0086} & {} & {} \\
    \midrule

    MoCA3D w/o box prior
                       & 16.35 & 10.88 & 0.3736 & 19.0 & 27.0 \\
    \quad $\Delta$ vs.\ MoCA3D
                       & \deltaCell{+0.29} & \deltaCell{+0.05} & \deltaCell{-0.0032} & {} & {} \\
    \midrule

    Direct Regressor   & 17.38 & 14.89 & 0.3064 & 22.1 & 19.2 \\
    \quad $\Delta$ vs.\ MoCA3D
                       & \deltaCell{+1.32} & \deltaCell{+4.06} & \deltaCell{-0.0704} & {} & {} \\
    \bottomrule
  \end{tabular}%
  }
  \caption{\textbf{MoCA3D ablations.}
  We report PAG$_{uv}$, PAG$_d$ (lower is better) and IoU$_{3D}$ (higher is better); $\Delta$ denotes \textit{method} $-$ MoCA3D.}
  \label{tab:pag_ablation}
\end{table}

We conduct controlled ablations on Objectron~\cite{objectron}, keeping the overall MoCA3D pipeline unchanged.
Table~\ref{tab:pag_ablation} reports PAG (positive delta indicates degradation), and IoU$_{3D}$ (negative delta indicates degradation). In addition, it reports the trainable parameter count and training compute for each variant.

\noindent\textbf{MoCA3D with Depth Anything~\cite{depthanything3}.} We augment MoCA3D with Depth Anything v3~\cite{depthanything3} depth cues by fusing mono and metric predictions into a feature stream alongside the DINOv3~\cite{simeoni2025dinov3} feature map.
This variant shows a marginal degradation on all metrics (+0.09px in PAG$_{uv}$, +0.09\% in PAG$_d$, and -0.0086 in IoU$_{3D}$), suggesting that MoCA3D's box-conditioned dense geometry learning already provides sufficiently strong cues, while injecting external depth features can introduce redundancy or noise.

\noindent\textbf{Removing box prior map.} Disabling box prior map conditioning degrades on all metrics (+0.29px in PAG$_{uv}$, +0.05\% in PAG$_d$, and -0.0032 in IoU$_{3D}$), indicating that explicit box-relative geometry cues are important for pixel-level corner localization.

\noindent\textbf{Direct regression head.} We replace the dense heatmap and depth heads with an RoI-aligned direct regressor that predicts corner $(u,v,d)$ values in a single shot. This yields the largest performance drop (+1.32px in PAG$_{uv}$, +4.06\% in PAG$_d$, and -0.0704 in IoU$_{3D}$), supporting our design choice that dense, pixel-aligned supervision is critical for stable optimization and precise image-plane geometry recovery.

\subsection{Downstream Task: Controllable Driving Scene Generation}

\begin{figure*}[t]
  \centering
  \includegraphics[height=7.5cm]{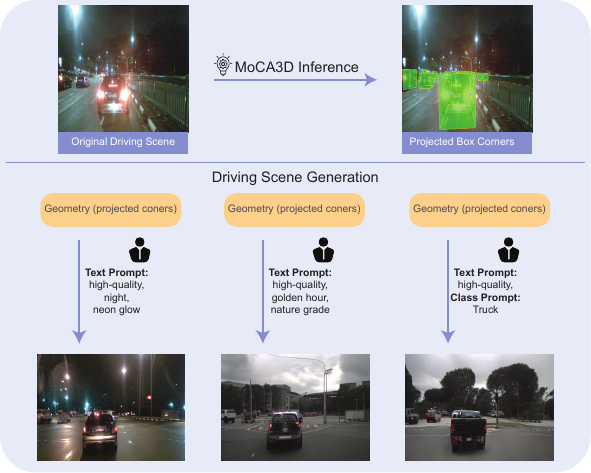}
  \caption{\textbf{Driving scene variation guided by MoCA3D.} MoCA3D recovers image-plane vehicle geometry (projected 3D box corners) from a single image and uses it to condition diffusion-based generation for diverse prompt-driven edits.
  }
  \label{fig:driving}
\end{figure*}

Recent controllable street-view synthesis methods~\cite{zhang2025perldiff, swerdlow2024street, yang2025instadrive, yang2023bevcontrol} rely on explicit 3D geometric controls (e.g., perspective-projected layouts). For instance, PerLDiff~\cite{zhang2025perldiff} leverages perspective projection cues as geometric priors and uses text prompt to manipulate lighting and weather conditions of street scenes.
However, these controls are typically sourced from 3D annotations, where projected 3D boxes provide eight 2D corner points per object.

We show that MoCA3D can replace this annotation-dependent geometric input.
Given a single real driving image, we run MoCA3D to obtain image-plane geometry for vehicles as projected cuboid corners.
We then use the recovered geometry as the object-level control signal in a diffusion-based generation pipeline, while changing the text and class prompt to specify the desired environmental condition.
As shown in Fig.~\ref{fig:driving}, this produces diverse driving-scene variations that preserve the underlying viewpoint and vehicle layout, demonstrating that image-plane geometry is sufficient to enable practical, annotation-free controllable generation from raw images.

\section{Conclusion and Future Work}
In this work, we introduce MoCA3D, a monocular, class-agnostic approach that represents object geometry directly in the image-plane as projected 3D cuboid corners with depths. By casting recovery as dense prediction, MoCA3D yields stable, accurate pixel-aligned geometry without relying on camera intrinsics.
We also introduce Pixel-Aligned Geometry (PAG) to directly evaluate corner and depth consistency in the image plane. MoCA3D achieves state-of-the-art PAG performance while remaining competitive on standard 3D metrics.

MoCA3D currently assumes that all eight corners are visible, and thus struggles with truncated objects. In addition, corner-only geometry leaves orientation ambiguity for symmetric instances. Future work could mitigate these issues via truncation-augmented preprocessing or by incorporating directional cues and auxiliary keypoints.

\section*{Acknowledgements}
\noindent\textbf{Icon credits.}
“Inference” icon by Wehape and “Man” icon by Viktor Vorobyev, from The Noun Project (CC BY 3.0).

%
%

\clearpage
\appendix
\renewcommand{\thesection}{\Alph{section}}
\renewcommand{\thesubsection}{\Alph{section}.\arabic{subsection}}

\section{Additional Method Details}
\subsection{Box Prior Encoding and Box-Conditioned Decoder}
We provide additional implementation details on how the oracle 2D box is injected into the transformer.
Given a normalized box $\mathbf{b}=(x_1, y_1, x_2, y_2)$, we first construct a dense four-channel prior map on the backbone feature grid.
For each feature location $(x, y)$, we compute box-relative center offsets and top-left-relative normalized coordinates:
\begin{equation}
d_x = \frac{x-c_x}{w_b}, \qquad
d_y = \frac{y-c_y}{h_b}, \qquad
u = \frac{x-x_1}{w_b}, \qquad
v = \frac{y-y_1}{h_b},
\end{equation}
where $(c_x, c_y)$ is the box center and $(w_b, h_b)$ denote the box width and height.
In the implementation, $(x,y)$ corresponds to the center of each backbone token on the normalized feature grid, i.e., $(x,y)\in((0,1),(0,1))$.
The resulting prior tensor is projected to the transformer hidden dimension using a lightweight $1\times1$-ReLU-$1\times1$ module, and then fused with the DINOv3~\cite{simeoni2025dinov3} feature map through a learnable gated residual~\cite{he2016deep} connection.

We also encode the input box as a set of box embeddings for decoder conditioning.
Rather than using a single box token, we form $9$ geometric keypoints from the box: the four corners, four edge midpoints, and the box center. Their normalized coordinates are transformed with a sine-cosine positional embedding, passed through an MLP, and augmented with learned identity embeddings so that each token preserves its geometric role.
These box tokens are then used as keys and values in decoder cross-attention.

Our decoder keeps dense image tokens $\mathbf{X}\in\mathbb{R}^{N\times C}$ as queries and treats box embeddings $\mathbf{B}\in\mathbb{R}^{N_q\times C}$ as conditioning keys/values:
\begin{equation}
\mathrm{Attn}(\mathbf{X},\mathbf{B})
=
\mathrm{softmax}\!\left(\frac{(\mathbf{X}W_Q)(\mathbf{B}W_K)^\top}{\sqrt{d}}\right)\mathbf{B}W_V,
\end{equation}
Since cross-attention preserves the query shape, the output remains a dense $N\times C$ token map, which is well suited for pixel-aligned prediction and resembles conditioned generation pipelines such as Stable Diffusion~\cite{rombach2022high}, where spatial tokens are modulated by external conditions through attention.

\subsection{Virtual Depth Parameterization}
\label{sec:virtual_depth}

Following Cube R-CNN~\cite{brazil2023omni3d}, we supervise depth in \emph{virtual depth} space rather than raw metric depth. This is important for Omni3D, which aggregates images captured with diverse cameras and focal lengths. Virtual depth reduces cross-camera scale variation by mapping metric depth into a normalized camera space with fixed virtual focal length $f_v$ and virtual image height $H_v$.

Given the metric depth $d_i$, image height $H$, and focal length $f$, its virtual depth target is defined as:
\begin{equation}
d_i^{v} = d_i \cdot \frac{f_v}{f}\cdot\frac{H}{H_v},
\end{equation}
where $f_v$ and $H_v$ are fixed hyperparameters ($f_v$=$H_v$=$512.0$ in our implementation) shared across the dataset. Since the conversion depends only on camera- and image-level quantities, all corners share the same scaling.

At inference, MoCA3D predicts virtual depth, which can be converted back to metric depth when camera intrinsics are available:
\begin{equation}
d_i = d_i^{v}\cdot\frac{f}{f_v}\cdot\frac{H_v}{H}.
\end{equation}
When camera intrinsics are unavailable at inference time, we assume a canonical focal length $f=f_v$.

\subsection{MoCA3D-Cube Adapter Details}
\begin{figure*}[!t]
  \centering
  \includegraphics[width=\textwidth]{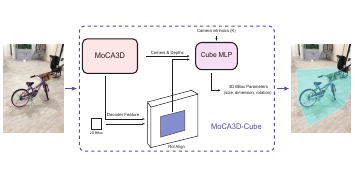}
  \caption{\textbf{MoCA3D-Cube Adapter.}
    MoCA3D-Cube combines MoCA3D's predicted corner-depth pairs with an RoI-aligned decoder feature and camera intrinsics $K$ to regress a parametric 3D bounding box. The adapter reuses image-plane geometry as the primary cue while adding a lightweight Cube MLP for conventional box prediction.}
  \label{fig:moca3dcube}
\end{figure*}

As illustrated in Fig.~\ref{fig:moca3dcube}, MoCA3D-Cube uses the MoCA3D corner and depth outputs to define geometric priors for box center and size, and the Cube MLP predicts residual updates on top of them. Specifically, the predicted corner-depth pairs are backprojected to 3D corners $(X_i, Y_i, z_i)$ using camera intrinsics, and their center prior is computed as the mean corner location:
\begin{equation}
    (X_0, Y_0, z_0)=\frac{1}{8}\sum_i (X_i, Y_i, z_i)
\end{equation}
The size prior is computed from the mean center-to-corner radius:
\begin{equation}
    r=\frac{1}{8}\sum_i\|(X_i, Y_i, z_i) - (X_0, Y_0,z_0)\|_2
\end{equation}
 This is converted into a cube prior $\mathbf{s}_0=(2/\sqrt{3})\,r\cdot (1,1,1)$.

The Cube MLP takes two inputs: a geometry feature, formed from per-corner offsets $(x_i^{\mathrm{norm}}-x_0,\; y_i^{\mathrm{norm}}-y_0,\; \log z_i-\log z_0)$ over the 8 corners together with the global prior term $(x_0, y_0, z_0, X_0, Y_0)$, and the RoI feature obtained by applying \texttt{roi\_align} with an output size $7\times7$ to the tight 2D box, followed by a $1\times1$ convolution, ReLU, and global average pooling:
\begin{equation}
\mathbf{r}=\mathrm{GAP}\!\left(\mathrm{ReLU}\!\left(\mathrm{Conv}_{1\times1}\!\left(\mathrm{RoIAlign}(\mathbf{F}_{\mathrm{dec}},\mathbf{b})\right)\right)\right),\qquad
\mathbf{o}=\mathrm{MLP}\!\left([\mathbf{g};\mathbf{r}]\right),
\end{equation}
where $\mathbf{o}\in\mathbb{R}^{13}$ denotes the predicted residual box parameters, rotation, and uncertainty.

\section{Data and Preprocessing}
\subsection{Omni3D and Dataset Notes}

\begin{table}[t]
\centering
\caption{Image statistics of the Omni3D source datasets.}
\label{tab:omni3d_source_stats}
\begin{tabular}{lrrrr}
\toprule
Dataset & Total & Train & Val & Test \\
\midrule
KITTI        & 7,481  & 3,321  & 391   & 3,769 \\
SUN RGB-D    & 10,335 & 4,929  & 356   & 5,050 \\
nuScenes     & 34,149 & 26,215 & 1,915 & 6,019 \\
Objectron    & 46,644 & 33,519 & 3,811 & 9,314 \\
ARKitScenes  & 60,924 & 48,046 & 5,268 & 7,610 \\
Hypersim     & 74,619 & 59,543 & 7,386 & 7,690 \\
\midrule
Omni3D       & 234,152 & 175,573 & 19,127 & 39,452 \\
\bottomrule
\end{tabular}
\end{table}

Our experiments build on Omni3D~\cite{brazil2023omni3d}, a large-scale benchmark for image-based 3D object understanding. After merging semantically overlapping categories across sources, Omni3D contains 234,152 images with roughly 3 million labeled 3D bounding boxes spanning 98 object categories. Table~\ref{tab:omni3d_source_stats} shows the detailed statistics of Omni3D.

All annotations are expressed in a unified camera-centric coordinate system with the camera center as origin and axes defined as +x right, +y down, and +z inward. Each instance is annotated with a category, a 2D box, a 3D centroid, a rotation matrix, and metric box dimensions. Input resolutions range from 370 to 1920 pixels and focal lengths from 518 to 1708 pixels.

\subsection{Preprocessing Procedure}

\subsubsection{2D Box completion with SAM2~\cite{ravi2024sam2} and Filtering Criteria}

Some Omni3D annotations do not provide a valid tight 2D bounding box. To make such instances usable in our box-conditioned setting, we first complete missing \texttt{bbox2D\_tight} annotations using SAM2. Concretely, for each object whose tight 2D box is missing, we construct a rough box from the min/max coordinates of the projected 3D cuboid corners and use it as a box prompt for SAM2. We then convert the predicted mask into a tight 2D bounding box by taking the extremal mask coordinates, and retain the update only when the predicted region is non-empty and spatially valid.

After box completion, we apply a two-stage procedure. First, we follow the standard Cube R-CNN annotation filtering. Specifically, this step removes ignored, invalid, or severely corrupted annotations according to dataset metadata and geometric validity checks; we refer readers to Omni3D~\cite{brazil2023omni3d} for the full filtering details.

Second, we apply an additional filter. We keep only annotations marked as \texttt{Good}, i.e., instances whose projected cuboid corners all lie inside the image, and further discard very small objects using a minimum resized box-area threshold (at least 1024 pixels$^2$ after resizing the longest image side to 512). We enforce the \texttt{Good} constraint because MoCA3D predicts image-plane corners with soft-argmax, making targets with out-of-image corners unsuitable for reliable supervision. After preprocessing, the final training set contains approximately 364K samples. The same filtering procedure is applied to the validation and test sets, and all experiments are conducted on this filtered subset.

\subsubsection{Image-Space Canonicalization and Input Preparation}

Because MoCA3D predicts geometry directly in the image plane, we canonicalize the target corner order using 2D image coordinates rather than a dataset-specific 3D vertex convention. For each instance, we reorder the eight projected cuboid corners by splitting them into the lower and upper four points according to their image-plane vertical coordinate, and then sorting each group from left to right using the horizontal coordinate. The same permutation is applied to the corresponding depth values, and this image-aligned ordering provides a more stable supervision format.

Our dataloader then converts each annotation into image-aligned training targets. Images are resized to $512\times 512$ with aspect-ratio-preserving letterboxing and normalized using ImageNet~\cite{5206848} mean and standard deviation. Projected corners and tight 2D boxes are transformed with the same scale and padding, and are normalized to $[0, 1]$. Corner depths are normalized to virtual depth (Sec.~\ref{sec:virtual_depth}).

\subsubsection{Augmentation}

\begin{figure*}[!t]
  \centering
  \includegraphics[width=\textwidth]{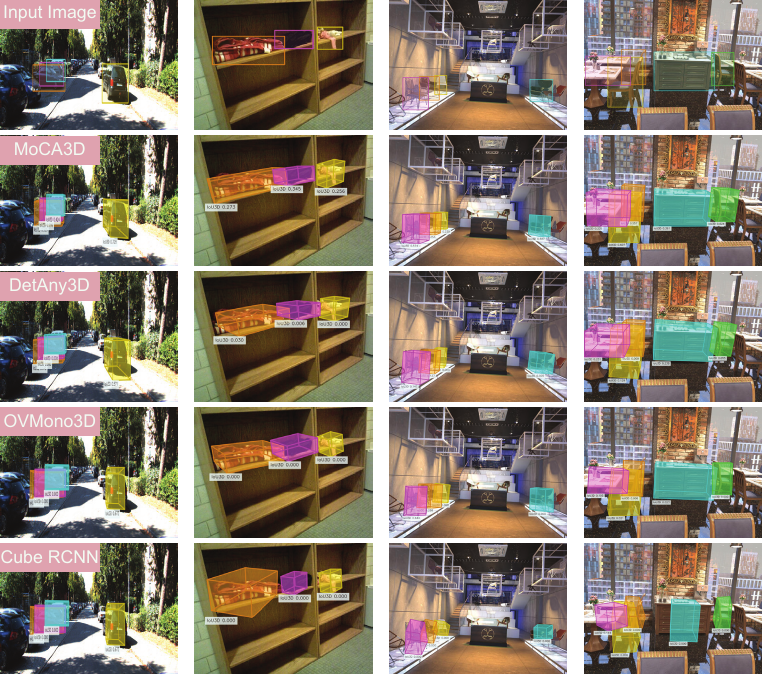}
  \caption{\textbf{Qualitative comparisons under oracle 2D boxes.}
    MoCA3D is compared with DetAny3D, OVMono3D, and Cube R-CNN on representative samples. Each prediction is shown together with its 3D IoU score, which is also displayed inside the corresponding image. Across diverse scenes, MoCA3D generally yields more accurate and visually consistent 3D geometry.}
  \label{fig:supp_qual}
\end{figure*}

In the feature-based training pipeline, we do not rely on photometric augmentation. Instead, augmentation is applied primarily in feature space and is synchronized with corresponding geometric targets. Specifically, during training we apply random horizontal flipping with probability 0.4 and random in-plane rotation within $\pm 10^\circ$ to the pre-extracted DINOv3 feature map. 

We further use lightweight stochastic perturbations in feature and box space. After a 5-epoch warm-up, Gaussian feature noise with standard deviation 0.01 is added to the DINOv3 feature map, and Gaussian jitter with standard deviation 0.02 is applied to the normalized 2D box coordinates.

In addition, we apply token-level regularization inside the model. During training, spatial feature tokens are randomly masked with probability 0.25, using either independent random masking or, with probability 0.4, block masking with block size $4\times 4$. When enabled, masked tokens are also excluded from transformer attention.

\section{Additional Experimental Results}
\subsection{Qualitative Comparisons}

Fig.~\ref{fig:supp_qual} presents qualitative comparisons under the oracle-2D setting on samples from Omni3D. We compare MoCA3D with DetAny3D~\cite{zhang2025detect}, OVMono3D~\cite{yao2024open}, and Cube R-CNN~\cite{brazil2023omni3d}. The predicted 3D boxes are visualized together with their 3D IoU scores, which are also shown inside each image for direct reference. Overall, MoCA3D produces more accurate and visually consistent projected cuboids, especially for challenging cross-domain cases involving scale variation, clutter, and indoor layouts.

\subsection{Driving Scene Generation Results}

\begin{figure*}[!t]
  \centering
  \includegraphics[width=\textwidth]{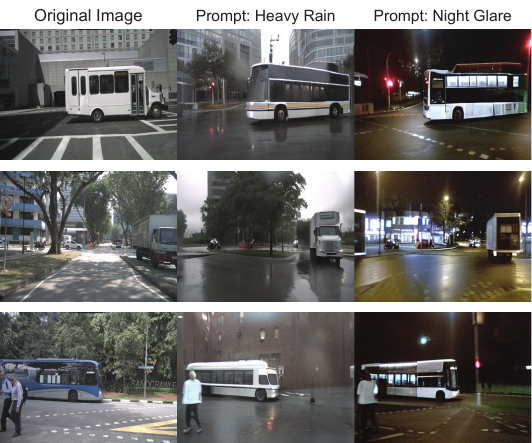}
  \caption{\textbf{Additional driving scene generation results.}
    Additional examples of controllable driving-scene editing using MoCA3D-recovered image-plane geometry.}
  \label{fig:supp_drive}
\end{figure*}

Fig.~\ref{fig:supp_drive} provides additional qualitative results for controllable driving-scene generation.
We follow the same pipeline used in the main paper, where MoCA3D first recovers image-plane vehicle geometry from a single real image, and the projected cuboid corners are then used as control signals in the generation pipeline.
By keeping the geometric control fixed while changing only the text prompt, the pipeline produces diverse scene variations such as heavy rain and night glare while preserving viewpoint, placement, and road layout. Even under substantial appearance changes, the results remain consistent with the source image.

\subsection{Failure Cases}

\begin{figure*}[!t]
  \centering
  \includegraphics[width=\textwidth]{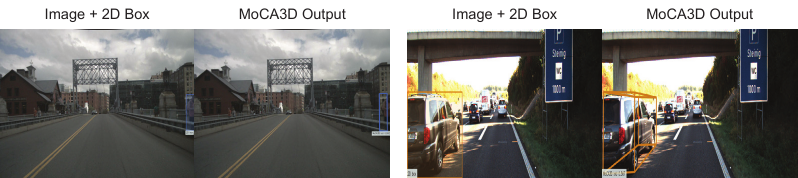}
  \caption{\textbf{Failure Cases.}
    We show representative failure cases of MoCA3D. For each example, the left panel shows the input image with the tight oracle 2D box, and the right panel shows the corresponding MoCA3D prediction.}
  \label{fig:fail}
\end{figure*}

Fig.~\ref{fig:fail} illustrates representative failure cases of MoCA3D.
A common failure mode arises when an object is truncated by the image boundary.
Since MoCA3D predicts pixel-aligned geometry directly in the image plane, its predictions are naturally constrained by the visible image extent.
As a result, when a portion of the object lies outside the frame, the predicted corners may collapse toward the boundary or yield inaccurate 3D geometry.

This behavior highlights a boundary case of image-plane geometry prediction rather than a failure of the overall representation.
Incorporating explicit reasoning for partially unobserved object extent is a promising direction for future work.

\section{Reproducibility Details}

All results in this appendix correspond to the same models, training protocol, and experimental setup described in the main paper.

\subsection{Implementation Details}

Unless otherwise noted, MoCA3D is trained in feature mode with pre-extracted DINOv3 features rather than end-to-end RGB backbone optimization. The model uses a DINOv3 image size of 512, predicts heatmaps at resolution $128\times128$, and applies soft-argmax with $\beta=100$. A $1\times1$ projection maps the 1024-channel DINOv3 feature to the transformer dimension $d=384$. Both the transformer encoder and decoder use 4 layers with 6 attention heads and feed-forward dimension 1024. The model predicts corner heatmaps and corner depth maps, and uses ReLU activation with dropout 0.15 throughout. The depth head uses a hidden dimension of 256, bottleneck dimension of 64, and minimum depth of $10^{-3}$. 

Training is performed for 120 epochs with 5 warm-up epochs, effective batch size 192, learning rate $2\times10^{-4}$, and weight decay 0.02. We use decoder and box-embedding learning-rate multipliers of 1.5 and 2.0, respectively. The training epoch length is fixed to 50,000 samples, validation is performed every 5 epochs, and checkpoints are saved every 60 epochs.

\subsection{Details of Ablation Models}

\subsubsection{MoCA3D with Depth-Anything}
This variant augments MoCA3D with depth predictions from Depth Anything3~\cite{depthanything3}, namely a monocular depth output and a metric depth output. All notations in this section follow the main paper.
Concretely, given an input image $I$, we first extract:
\begin{equation}
\mathbf{D}^{\text{mono}} = f_{\text{mono}}(I), \qquad
\mathbf{D}^{\text{metric}} = f_{\text{metric}}(I),
\end{equation}
where $f_{\text{mono}}, f_{\text{metric}}$ denote the two Depth-Anything branches, and the two depth outputs are embedded by a dedicated depth embedder:
\begin{equation}
\mathbf{E}^{\text{da}} = \phi_{\text{da}}(\mathbf{D}^{\text{mono}}, \mathbf{D}^{\text{metric}}),
\end{equation}
where $\mathbf{E}^{\text{da}}\in\mathbb{R}^{C\times h_d\times w_d}$.
Additionally, we inject the box prior map into $\mathbf{E}^{\text{da}}$:
\begin{equation}
\tilde{\mathbf{E}}^{\text{da}} = \mathbf{E}^{\text{da}} + \alpha \mathbf{E}_{\text{prior}},
\end{equation}
so that both the appearance stream and the auxiliary depth stream are aligned to the same box-relative geometry.

Unlike the baseline MoCA3D, this variant replaces the encoder with a depth-aware fusion module that integrates appearance and auxiliary depth tokens. After adding standard 2D positional encodings and flattening both $\tilde{\mathbf{F}}$ and $\tilde{\mathbf{E}}^{\text{da}}$ into token sequences, the module applies a decoder-like layer structure consisting of self-attention, cross-attention, and a feed-forward network. In particular, the DINOv3 token stream serves as the query sequence, while the depth-aware tokens are used as keys and values in cross-attention.
The fused token sequence is then reshaped back to the feature grid and passed to the same decoder and dense heads as in MoCA3D. Therefore, this ablation modifies only the feature fusion stage, enabling a controlled comparison of whether auxiliary depth cues improve geometry recovery.

\subsubsection{Direct Regression Model}

This ablation replaces MoCA3D's dense upsampling and prediction heads with a direct regression head. Instead of predicting eight heatmaps and eight depth maps, the direct model first fuses the decoder feature and the skip feature by channel-wise concatenation followed by a fusion block. Given the fused feature map, we apply RoI-Align to the input 2D box with output size of $7\times7$. The cropped feature is then channel-wise compressed, flattened, and passed through a two-layer MLP to directly regress an output vector.

The output is interpreted as the image-plane 3D geometry itself: the first $16$ values correspond to the projected eight corners, and the remaining $8$ values correspond to the assigned depths. Compared with MoCA3D, this variant removes the dense pixel-aligned formulation, and the task is reduced to an RoI-to-vector regression, making this ablation a clean control experiment for testing whether MoCA3D's gains arise from its dense geometry prediction design.

\bibliographystyle{splncs04}
\bibliography{main}
\end{document}